\documentclass[a4paper,conference]{IEEEtran}

\usepackage[pdftex]{graphicx}

\usepackage{amsmath, amssymb}
\newtheorem{proposition}{Proposition}

\usepackage{graphicx}
\usepackage{amsmath}
\usepackage{amssymb}
\usepackage{bm}
\usepackage[T1]{fontenc}
\usepackage[ruled,vlined]{algorithm2e}
\usepackage{color}

\begin{document}
\title{Variational Auto Encoder Gradient Clustering}

\author{\IEEEauthorblockN{Adam Lindhe}
\IEEEauthorblockA{Department of Mathematics\\
KTH Royal Institute of Technology\\
100 44 Stockholm, SWEDEN\\
Email: adlindhe@kth.se}
\and
\and
\IEEEauthorblockN{Carl Ringqvist}
\IEEEauthorblockA{Department of Mathematics\\
KTH Royal Institute of Technology\\
100 44 Stockholm, SWEDEN\\
Email: carrin@kth.se}
\and
\IEEEauthorblockN{Henrik Hult}
\IEEEauthorblockA{Department of Mathematics\\
KTH Royal Institute of Technology\\
100 44 Stockholm, SWEDEN\\
Email: hult@kth.se}
}

\maketitle

\begin{abstract}
Clustering using deep neural network models have been extensively studied in recent years. Among the most popular frameworks are the VAE and GAN frameworks, which learns latent feature representations of data through encoder / decoder neural net structures. This is a suitable base for clustering tasks, as the latent space often seems to effectively capture the inherent essence of data, simplifying its manifold and reducing noise. In this article, the VAE framework is used to investigate how probability function gradient ascent over data points can be used to process data in order to achieve better clustering. Improvements in classification is observed comparing with unprocessed data, although state of the art results are not obtained. Processing data with gradient descent however results in more distinct cluster separation, making it simpler to investigate suitable hyper parameter settings such as the number of clusters. We propose a simple yet effective method for investigating suitable number of clusters for data, based on the DBSCAN clustering algorithm, and demonstrate that cluster number determination is facilitated with gradient processing. As an additional curiosity, we find that our baseline model used for comparison; a GMM on a t-SNE latent space for a VAE structure with weight one on reconstruction during training (auto encoder), yield state of the art results on the MNIST data, to our knowledge not beaten by any other existing model.
\end{abstract}

\IEEEpeerreviewmaketitle

\section{Introduction}

Consider a data set containing subsets with several distinct features. Clustering algorithms aim at partitioning the data into groups with similar features. The number of clusters is a hyper parameter that is generally unknown and determining its value is crucial for successful clustering. 

An approach to determine the number of clusters is to apply a given clustering algorithm over a range of values of the number of clusters and find the number that minimize some cost function, such as the within-cluster sum of squares. Such methods include the 
CH index (Calinski and Harabasz 1974)
Hartigan’s rule (Hartigan 1975), the silhouette statistic (Kaufman and Rousseeuw 1990), approximate Bayes factors (Kass and Raftery 1995; Frayley and Raftery 1998), the gap statistic (Tibshirani, Walther, and Hastie 2001), the information theoretic approach by Sugar and James (2003), and the lower-bound criterion (Steinley and Brusco, 2011).  

Akaike's information criterion for estimating the number of clusters is used in e.g.\ Bozdogan and Sclove (1984), where the Bayesian information criterion BIC  Schwarz et al. (1978) is  Ishioka (2005), Zhao, Hautamaki, and Fränti (2008) and Cheong and Lee (2008). 

Consensus or ensemble methods aim at combining algorithms to find the robust estimates on the number of clusters, see  Milligan and Cooper (1985)  (Dimitriadou, Dolničar, \& Weingessel, 2002), \cite{Unlu2019},
(Lancichinetti, Fortunato, 2012, Strehl, Ghosh, 2003, Topchy, Jain, Punch, 2005).
Dörnfelder, Guo, Komusiewicz, and Weller (2014).

Algorithms where the number of clusters is an outcome of the procedure include correlation clustering (Bansal, Blum, \& Chawla, 2004), AutoSOM (Newman \& Cooper, 2010) and cross-clustering (Tellaroli, Bazzi, Donato, Brazzale, and Draghici, 2016).

A natural approach to clustering is to use a multi-modal probability density $p(x)$ to model the data generating mechanism. The popular $k$-means algorithm can be viewed as an implementation of the EM-algorithm to maximize the likelihood of a mixture of standard normal distributions that models the data. Extensions include the Gaussian mixture model (GMM) that allows for general Gaussian mixtures. 

In this article, the multi-modality of the probability density defined by a VAE architecture is leveraged for clustering, using modes as cluster labels. In order to assign a mode to a data point $x$, the gradient of $p(x)$ is followed until it reaches a local maxima of $p(x)$. An importance sampling scheme is identified that 
greatly simplifies the estimation of gradient of the density of a VAE with normal prior. 

Note that the number of clusters here is only limited by the VAE architecture. However, no two data points will end up in the exact same point after gradient application. Thus, in order to count clusters, it is not enough to simply count the number of identical points after gradient processing. Nor is it possible to assign a cluster label to a new data point through simply following the gradient and collect the label from a dictionary of unique end-points. Rather, gradient processing results in all data points belonging to the same mode (or cluster) being "close", but not identical. Thus, the suggested algorithm constitute an effective method for processing data for subsequent application of simpler clustering algorithms such as K-means or GMM. Moreover, the processed data is suitable for inspection when it comes to determining a natural cluster number size, since the number of clusters needs not to be set for the VAE training nor for the gradient processing; and since gradient processing results in more prominent cluster separation. To this end, we present an algorithm for suggesting a suitable number of clusters, based on the DBSCAN method \cite{Ester1996}; resembling to a great extent methods within persistent homology. It is demonstrated empirically that the gradient processing together with the DBSCAN method are valuable tools for determining a suitable number of clusters for data.

\section{Related work} \label{relatedwork}
The general idea of letting modes of a probability density function represent cluster labels is not new. In \cite{Fukunga1975}, the general concept was introduced through the Mean Shift method. The concept was later developed in \cite{Kulczycki2010}, where mathematical properties of the method was investigated for a variety of applications. Although the mean shift is widely used, a complete proof for its convergence is still not known in general \cite{Ghassabeh2015}, although convergence has been proved under certain strict conditions \cite{Ghassabeh2013}. In this article, the concept is applied to the VAE framework \cite{Kingma2013}; demonstrating a simple gradient form obtained through an importance sampling step; where convergence is assured.

\noindent Finding suitable data representations for clustering tasks, not least through neural network architectures, has been extensively researched. In \cite{Xie2016}, feature representations and cluster assignments are simultaneously learned through applying a neural network map between data space and a latent space. The approach results in impressive classification performance, although the setup does not extend well to other tasks, such as image generation. In \cite{Jiang2017}, the VAE generative procedure is modelled by a GMM model and a neural network, essentially creating a VAE with multimodal latent space; with latent representations highly suitable for clustering. The GAN framework \cite{Goodfellow2014} is notably used with great success in \cite{Makhzani2017} and \cite{Zhao2019}. In \cite{Makhzani2017}, a generative PixelCNN is combined with a GAN inference network, resulting in a model that can impose arbitrary priors on the latent space. In \cite{Zhao2019}, the GAN framework is embedded into the EM algorithm to do clustering, semi-supervised classification and dimensionality reduction. Although impressive state-of-the-art results are achieved for clustering tasks with mentioned methods, they require the number of clusters to be chosen before training, rendering experiments on this hyper parameter quite burdensome. The method presented in this paper does not require clusters to be chosen before network training, and it is demonstrated that it in fact can be suitable for investigating suitable choices of this hyper parameter.
The generative frameworks of VAE and GAN are generally among the most popular frameworks for learning useful data representations for clustering and related tasks, and they have been deployed in a variety of extensions and mixtures in literature, see e.g \cite{Salimans2016, Makhzani2016, Maaloe2016, Abbasnejad2016, Kingma2014B, Nalisnick2016}. 

\section{The vae gradient and convolution}
\noindent The VAE decoder defines a probability distribution over the data space $p(x)$ through:

\begin{align*}
    p(x) = \int p(x|z) p(z) dz
\end{align*}

\noindent The gradient can thus be estimated through Monte Carlo simulation on form 

\begin{align*}
    \nabla p(x) = \frac{1}{n} \sum \nabla p(x|z_{n}); \quad z_{n} \sim p(z)
\end{align*}

\noindent However, $p(x|z)$ typically has mass only in a small neighborhood of $z$, why the simulation becomes ineffective. Thus it is suitable to introduce an importance sampling element. If considering the log gradient, it is possible to conclude the following.

\hspace{1pt}

\begin{proposition}
\label{prop1}
If $p(x|z)$ is normally distributed, it holds that 
\begin{align*}
 \nabla \mathrm{log} p(x) \propto \mathrm{E}_{p(z|x)}[\mu(z)] - x
\end{align*}
 where $\mu(z)$ is the expectation of $p(x|z)$.
\end{proposition}

\hspace{1pt}

{\it Proof:} It holds that 

\begin{align*}
    \nabla \mathrm{log} p(x) &= \frac{1}{p(x)}\int \nabla p(x|z)p(z) dz \\
    &= \frac{1}{p(x)} \int \nabla p(x|z) p(z) \frac{p(z|x)}{p(z|x)}dz \\
    &= \frac{1}{p(x)} \int \nabla p(x|z) p(z) \frac{p(z|x)}{p(x|z)p(z)/p(x)}dz \\
    &= \int \frac{\nabla p(x|z)}{p(x|z)} p(z|x)dz \\
    &\propto \int (\mu(z) - x) p(z|x)dz \\
    &= \mathrm{E}_{p(z|x)}[\mu(z)] - x
\end{align*}

\noindent The second to last step follows from the fact that $p(x|z)$ is normally distributed with mean $\mu(z)$. This completes the proof.

\hspace{1pt}

\noindent Note that proposition \ref{prop1} implies that local maxima are located at exactly the points where reconstruction is perfect (on average). In turn, this suggests that an expressive VAE architecture with good reconstruction would tend to have many local maxima, numbering in the order of data observations. It is therefore suitable in general to "smooth" the density, in order to achieve fewer local maxima and thus fewer clusters. The following smoothed version of $\mathrm{log}p(x)$ is thus considered:

\begin{align*}
    \mathrm{log}p_{\sigma}(x) = \int \mathrm{log}p(x - y) g(y) dy
\end{align*}

\noindent where $g(y)$ is a normal distribution of variance $\sigma$ and 0 expected value. The greater the variance, the greater the smoothing effect. Note that 

\begin{align*}
    \nabla \int \mathrm{log}p(x - y) & g(y) dy = \int \nabla \mathrm{log}p(x - y) g(y) dy \\
    &\propto \int [ \mathrm{E}_{p(z|x-y)}[\mu(z)] - x + y) ] g(y) dy \\
    &= \Big(\int y g(y)dy= 0, \int xg(y)dy = x \Big) \\
    &= \int \mathrm{E}_{p(z|x-y)}[\mu(z)] g(y) dy - x \\
    &= \mathrm{E}_{g(y)}[\mathrm{E}_{p(z|x-y)}[\mu(z)]] - x
\end{align*}

\noindent An estimate of the gradient can thus be obtained by monte carlo sampling

\begin{align*}
    \nabla \mathrm{log} p_{\sigma}(x) / c \approx \frac{1}{m} \sum_{k=0}^{m} \frac{1}{n} \sum_{i=0}^{n} \mu(z_{i}) - x
\end{align*}

\noindent where $z_{i} \sim p(z|x+\epsilon_{k}), \epsilon_{k} \sim N(0, \sigma)$.

\section{Algorithm choosing suitable number of clusters}
\noindent Many common clustering algorithms, e.g K-means and GMM, require the number of clusters to be set as a hyper parameter. For complex data sets, determining a suitable number of clusters is challenging. Therefore, extensive analysis is often needed in order to determine the very general shape of data, in order to set a suitable number of clusters. Persistent homology is a tool commonly applied for this problem. Here, each data observation is assigned a ball of radius $\epsilon$. If the balls of two data-points intersect, they are considered to belong to the same cluster. If $\epsilon$ is zero, each data point will be assigned its own cluster, while there always exists an $\epsilon$ for which all data is considered to belong to the same cluster. By varying the size of $\epsilon$ from zero until the number of clusters becomes one, and counting the resulting number of clusters for each epsilon, it is possible to inspect which number of clusters that seems to persist over many consecutive values of $\epsilon$. A persistent number over many epsilon is suggested as a suitable alternative for the number of inherent clusters of the data set. The DBSCAN clustering algorithm shares many similarities with this approach. Similarly, DBSCAN is essentially counting the number of connected components, but with an additional useful concept of \textit{core points}. A data point is said to be a core point if at least $m$ other points are within distance $\epsilon$ from it. Any point that is not within distance $\epsilon$ from a core point is considered noise, and are assumed to not belong to any cluster. Thus, with $m=1$, the algorithm is equivalent to the basic persistent homology approach.

\begin{figure}[!t]
 \centering
\includegraphics[width=260pt, height=200pt]{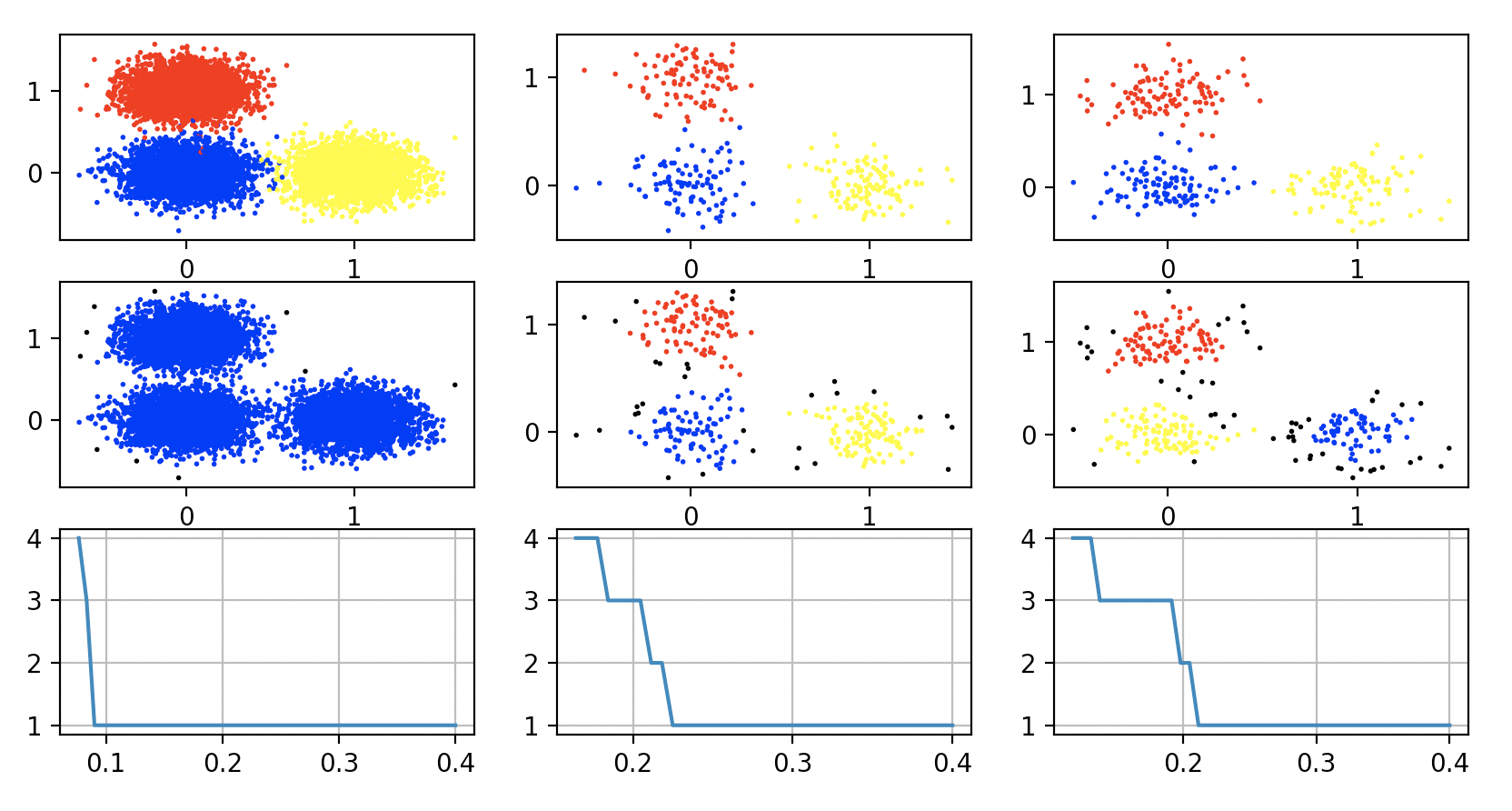}
\caption{\label{figure1} Visualisation of the advantage with sub-sampling when counting connected components with DBSCAN. Top left is 10000 samples from a mixed normal, colored with mode label. Top middle and top right are two sub-samples of size 300 colored with mode labels. Middle row is the result of DBSCAN on the top row data ($m=2, \epsilon=0.2$) with biggest three clusters colored red, blue and yellow, and remaining clusters and noise colored in black. The number of connected components are correctly identified for the sub-samples, but not for the whole sample. Bottom row shows number of clusters as chosen by DBSCAN over ball radius ($m=2$); for the whole data set bottom left, and for the respective sub samples bottom middle and bottom right. Noise label is disregarded. Persistence over 3 clusters is more prominent for the sub samples.}
\end{figure}

\noindent By computing the DBSCAN algorithm for fixed $m$ over varying $\epsilon$, and inspecting persistent number of clusters, a suitable suggestion of the number of clusters inherent in data can be formed. Here, a subset of data is sampled before computing persistence. This has the benefit of not only being computationally more efficient, but might emphasise cluster constitutions. For example, consider a data material consisting of 10000 samples from a mixed normal distribution, where modes are situated fairly close, so that intersection of data occurs but is not that common. Counting the number of connected components is problematic in this case, since paths of close points are easily generated between clusters. Since outliers of a distribution becomes less frequently observed with fewer samples, sub-sampling data reduces the probability of connected paths between clusters, as seen in Figure $\ref{figure1}$. This makes identification of the number of clusters through the DBSCAN approach simpler. In order to make identification more robust, the average over many simulations is chosen.

\section{Experiments on MNIST}
\noindent In this section, the gradient pre-processing of data and cluster counting algorithm is applied to the MNIST data set. Clustering accuracy is measured with the unsupervised clustering accuracy (ACC) metric \cite{Xie2016}

\begin{align*}
    ACC = \underset{m \in M}{\max} \frac{\sum_{i=1}^{N}1\{l_{i}=m(c_{i})\}}
    {N}
\end{align*}

\noindent where $l_{i}$ is the true label, $c_{i}$ is the assigned label, M is the set of all possible one-to-one mappings between assignments and labels, and N is the number of samples. The metric is essentially describing a scheme to map predicted labels to real labels, then computing the percentage of accurately predicted labels.

\begin{figure}[!t]
 \centering
\includegraphics[width=260pt, height=200pt]{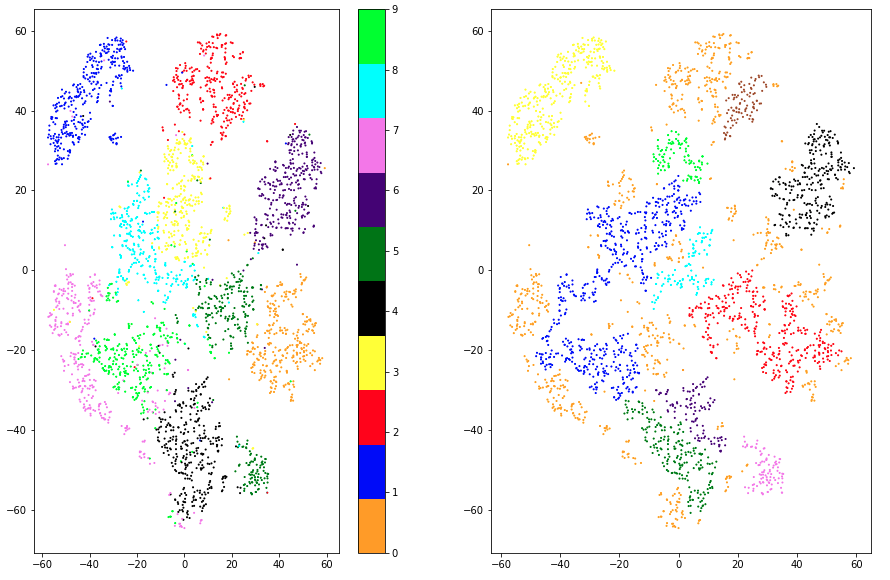}
\caption{\label{mnistVaeBeforeGradient} Visualisation over true labels (left) and DBSCAN clustering labels (right) in the t-SNE VAE latent space, MNIST data. Orange color in right picture is DBSCAN noise plus observations not having predicted label among the 10 most frequent labels.}
\end{figure}

\begin{figure}[!t]
 \centering
\includegraphics[width=170pt, height=170pt]{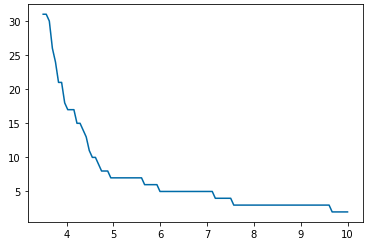}
\caption{\label{mnistVaePersistanceBeforeGradient} Sample persistance graph over t-SNE VAE latent space, MNIST data.}
\end{figure}

\noindent The MNIST data set \cite{LeCun1998} consists of white hand-written digits on black background. We use a modified version of the convolutional VAE architecture for MNIST as described in \cite{Berthelot2019}. For full description, see code \cite{CarlAdamHenrikCode}. In the experiments, the latent space has dimension 64. In order to facilitate visualisation and computation, the t-SNE-method \cite{Maaten2008} is used for projecting the latent space down to two dimensions. The VAE is trained over 1000 epochs with batch size 100, with 0.5 in weight on reconstruction and 0.5 in weight on the Kullback-Liebler (KL) distance. We use the Adam gradient optimizer with learning rate $10^{-4}$. The resulting latent space separation and DBSCAN classification over 3000 randomly chosen data points is displayed in Figure \ref{mnistVaeBeforeGradient}.
Although labels are not particularly mixed, they are not separated by space, making cluster constitution hard to detect without label colors (left in Figure \ref{mnistVaeBeforeGradient}). This shows also in the classification with DBSCAN (right in Figure \ref{mnistVaeBeforeGradient}). Here epsilon is $\epsilon=2$, $m=2$; and the 10 most frequent clusters are given a colour, while noise and remaining clusters are labelled 0, and given identical color orange. As can be seen, classification is poor, with the ACC measure being only 59.4 \%. The DBSCAN cluster counting gives unsatisfactory results as well. The average most persistent number of clusters over 50 runs, with 1000 as sub-sample size, yields 2.38 as result, far from the desired number 10 (number of digits). In Figure \ref{mnistVaePersistanceBeforeGradient} a sample curve is shown, indeed indicating (quite weak) signals for only low number of clusters.

\begin{figure}[!t]
 \centering
\includegraphics[width=250pt, height=200pt]{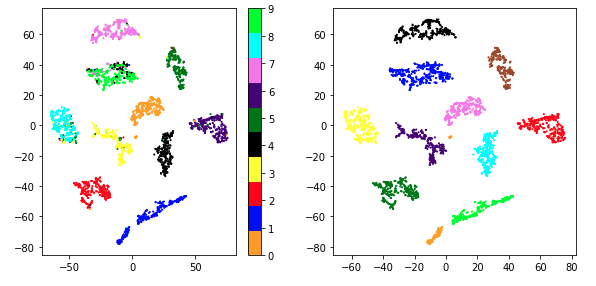}
\caption{\label{mnistVaeAfterGradient} Visualisation over true labels (left) and DBSCAN clustering labels (right) in the t-SNE VAE latent space after gradient processing, MNIST data.}
\end{figure}

\begin{figure}[!t]
 \centering
\includegraphics[width=180pt, height=180pt]{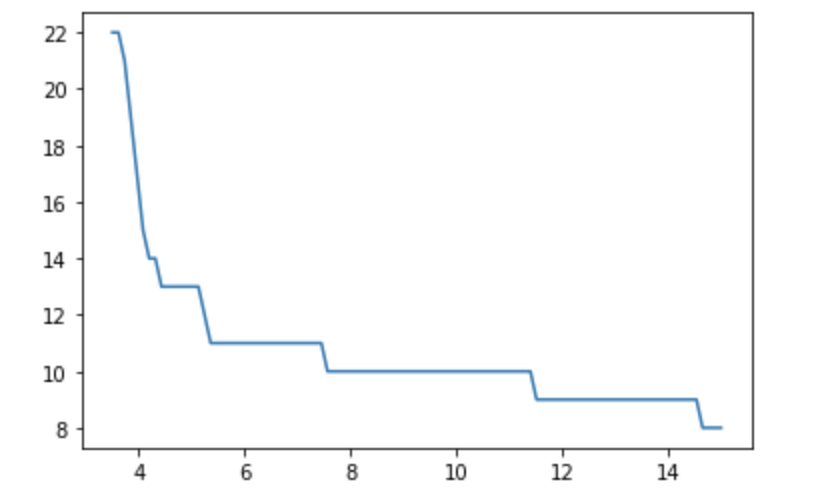}
\caption{\label{mnistVaePersistanceAfterGradient} Sample persistence graph over t-SNE VAE latent space, after gradient processing, MNIST data.}
\end{figure}

We then apply gradient steps as described to data. Again, we use a subset of 3000 samples. A gradient step of 0.001 is used over 7000 iterations, with smoothing variance $\sigma=0.0005$. The result can be seen in Figure \ref{mnistVaeAfterGradient}. Gradient processing has clearly resulted in clear and well separated clusters, with only minor errors in terms of mixed labels (left Figure \ref{mnistVaeAfterGradient}). This shows also in the classification with DBSCAN (right in Figure \ref{mnistVaeAfterGradient}). Here, $\epsilon=2$, $m=2$; and the 10 most frequent clusters are given a colour, while noise and remaining clusters are labelled 0, and given identical color orange. As can be seen, classification is significantly better, with the ACC now being 89.1 \%. The DBSCAN cluster counting is improved as well. Averaging the most persistent number of cluster over 50 runs yields 10.2, a near perfect result. In Figure \ref{mnistVaePersistanceAfterGradient} a sample curve is shown, indeed indicating strongest signal for cluster number 10.

\begin{figure}[!t]
 \centering
\includegraphics[width=200pt, height=200pt]{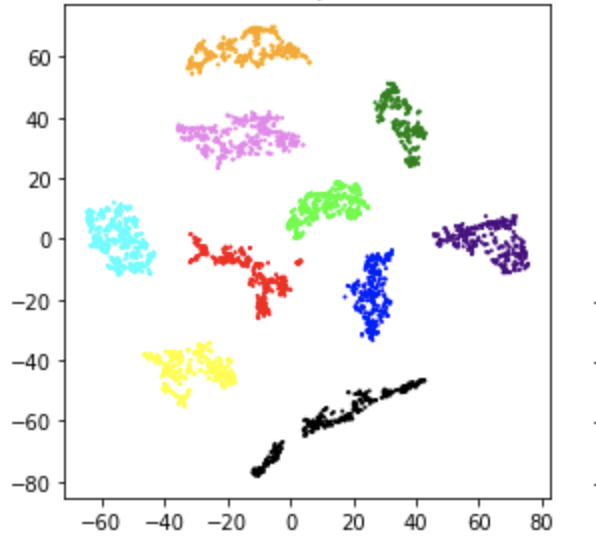}
\caption{\label{mnistVaeGmmAfterGradient} Visualisation over GMM clustering labels in the t-SNE VAE latent space after gradient processing, MNIST data.}
\end{figure}

\begin{figure}[!t]
 \centering
\includegraphics[width=250pt, height=180pt]{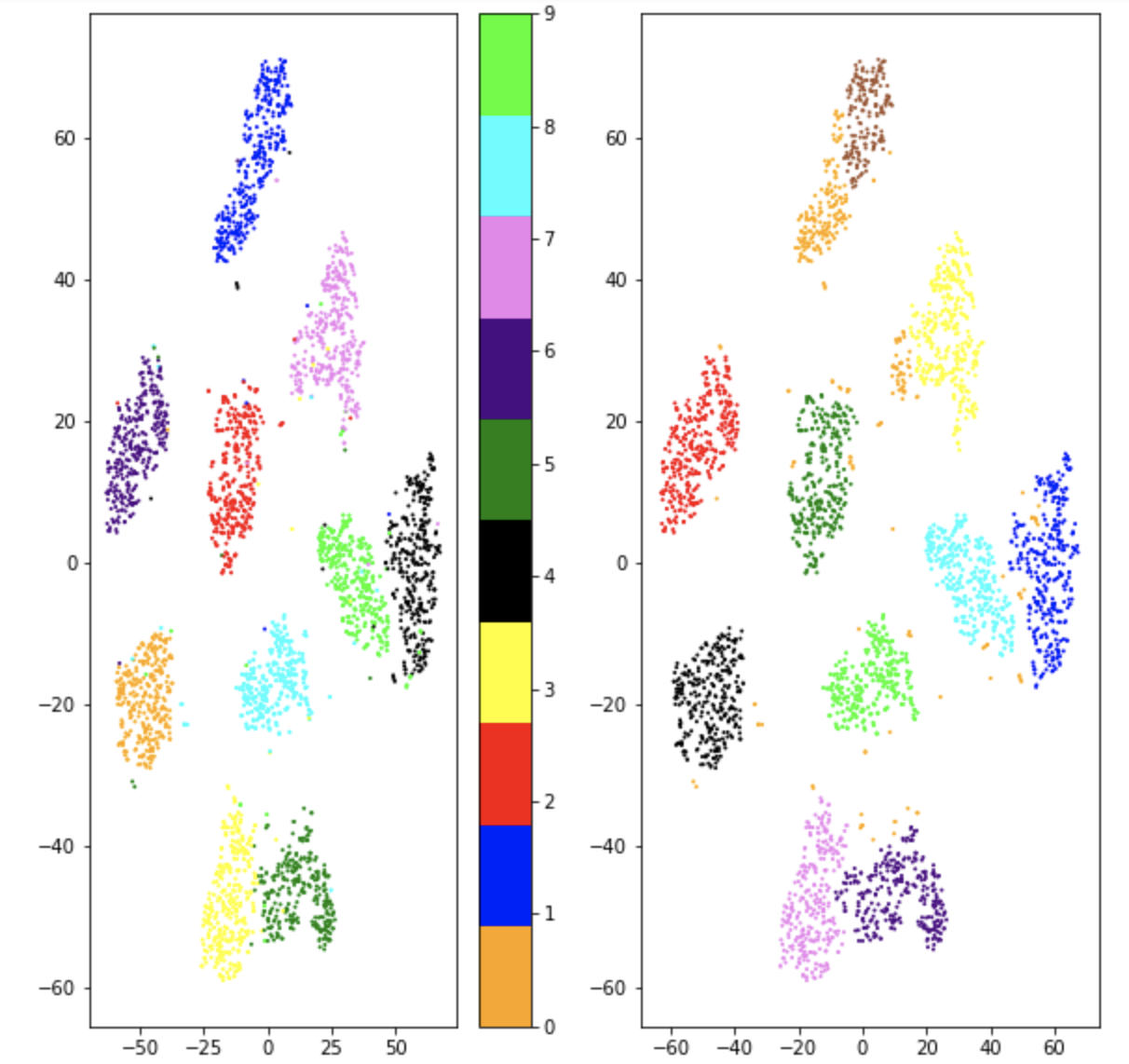}
\caption{\label{mnistAutoEncoderDbscan} Visualisation over true labels (left) and DBSCAN clustering labels (right) in the t-SNE VAE latent space, without gradient processing and weight one on reconstruction during training, MNIST data.}
\end{figure}

\begin{figure}[!t]
 \centering
\includegraphics[width=200pt, height=210pt]{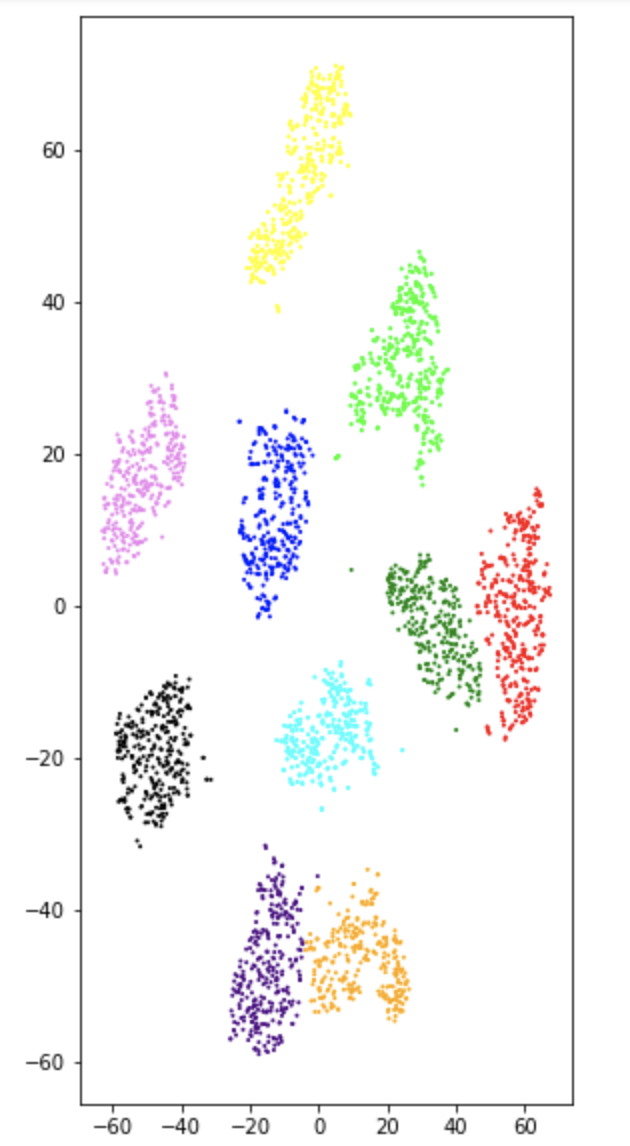}
\caption{\label{mnistAutoEncoderGmm}  Visualisation over GMM clustering labels in the t-SNE VAE latent space, without gradient processing and weight one on reconstruction during training, MNIST data.}
\end{figure}

\begin{figure}[!t]
 \centering
\includegraphics[width=180pt, height=180pt]{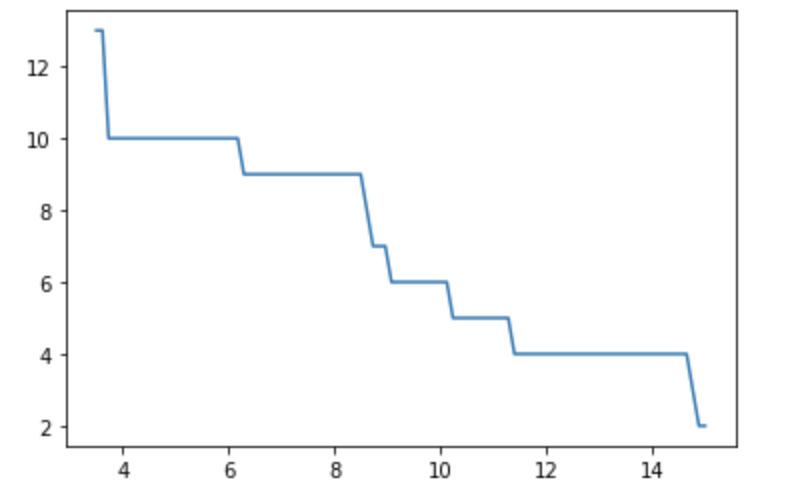}
\caption{\label{persistanceAutoencoder}  Sample persistence graph over t-SNE VAE latent space, without gradient processing and weight one on reconstruction during training, MNIST data.}
\end{figure}

As is seen in the Figures, DBSCAN fails to recognise the complete cluster for digit 1. This is the result of that DBSCAN essentially focuses on connected components, ignoring other important patterns, such as cluster variance structure. In Figure \ref{mnistVaeGmmAfterGradient}, the t-SNE VAE latent representations are instead classified with GMM. This improves the cluster accuracy to 92.3\%, approaching state of the art results; as it more correctly captures the variance structure of the clusters. The remaining error is the result of that the gradient processing is mixing labels. In turn, this could be the result of digit images looking similar to other digits (e.g an image of digit 4 looking similar to digit 9). This error type is hard to redeem, as it is inherent in data (even for humans, it is hard to determine labels for some MNIST digits). However, mixing might also occur due to bad reconstruction. Since the gradient processing is essentially moving images towards its reconstructed versions, bad reconstructions can result in image transitions towards wrong clusters. Since the VAE with its KL loss element is essentially sacrificing separation in latent space and reconstruction ability to the benefit of easy image generation; it is interesting to compare our approach with a pure autoencoder structure that are better preconditioned for prominent separation and accurate reconstruction to begin with. We therefore train an identical setup (architecture and hyper parameters) with weight one on reconstruction. Note that this will result in a very high number of local maximum in the probability function, why gradient processing is hard; substantial number of experiments confirm that gradient processing in this context results in near identical disposition of points for a variety of hyper parameters. When a change in disposition is detected, the result is typically unstable, with high mixing as result. The VAE trained with KL loss together with gradient processing will be referred to as VAE/KL/GRAD. The result in terms of separation and classification, without gradient processing, is seen in Figure \ref{mnistAutoEncoderDbscan}. To the left is the t-SNE latent data representations with true code label, and to the right is the result of DBSCAN. As can be seen, mixing between labels is nearly non-existent. The clustering accuracy for DBSCAN is however on par with DBSCAN in VAE/KL/GRAD, measuring 88.4\%. Although mixing is lower, collection within clusters and separation between clusters are both weaker; why DBSCAN has harder to identify all clusters correctly. Clear intra-cluster variance structures are although apparent, and applying GMM with 10 clusters results in a cluster accuracy of 97.14 \%, Figure \ref{mnistAutoEncoderGmm}; significantly better than for VAE/KL/GRAD. The DBSCAN cluster counting is however suffering, as seen in Figure \ref{persistanceAutoencoder}, since clusters are not as well separated as in VAE/KL/GRAD. Averaging the most persistent number of clusters over 50 runs yields the number 5.06, substantially worse than VAE/KL/GRAD. Thus, although our method of gradient processing does not seem to beat the pure autoencoder structure in terms of cluster accuracy, it helps substantially in determining the number of inherent clusters of data. Gradient processing also substantially improves cluster accuracy for a VAE with KL loss compared to when no gradient processing is applied to the same. 

It is interesting to note that our accuracy of 97.14\% is to our knowledge better than all state of the art methods of today on the MNIST data set, when using 10 clusters as hyper parameter. For example, in \cite{Jiang2017}, a cluster accuracy 94.46 \% is achieved, and in \cite{Zhao2019}, an accuracy of 95.8 \% is achieved.

\section{Experiments on Fashion-MNIST}
\begin{figure}[!t]
 \centering
\includegraphics[width=250pt, height=200pt]{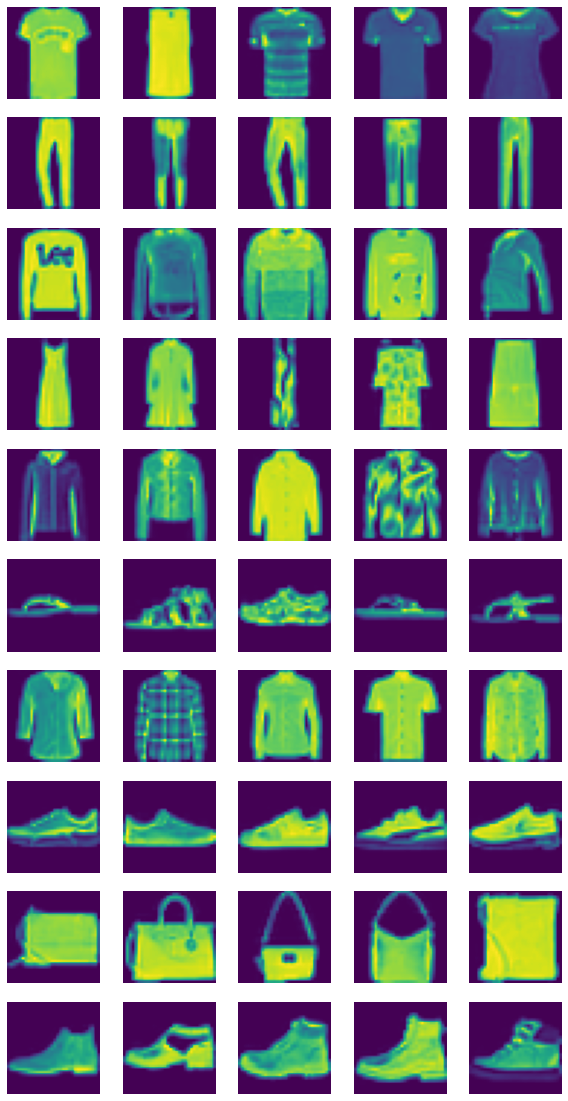}
\caption{\label{FmnistExamples} Examples from each true class from the Fashion-MNIST dataset}
\end{figure}

\begin{figure}[!t]
 \centering
\includegraphics[width=250pt, height=200pt]{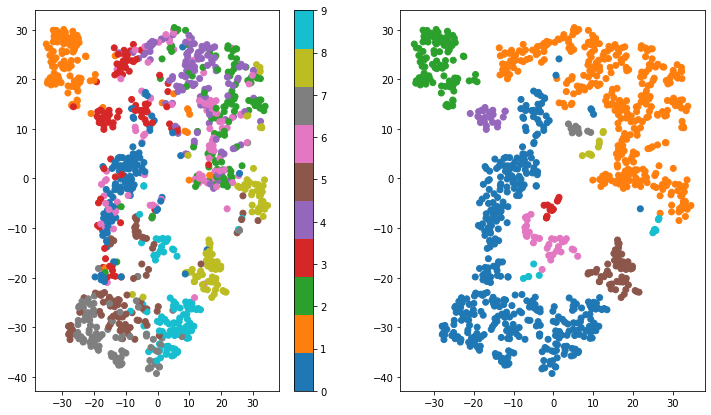}
\caption{\label{FmnistVaeBeforeGradient} Visualisation over true labels (left) and DBSCAN clustering labels (right) in the t-SNE VAE latent space, Fashion-MNIST data.}
\end{figure}

\begin{figure}[!t]
 \centering
\includegraphics[width=250pt, height=200pt]{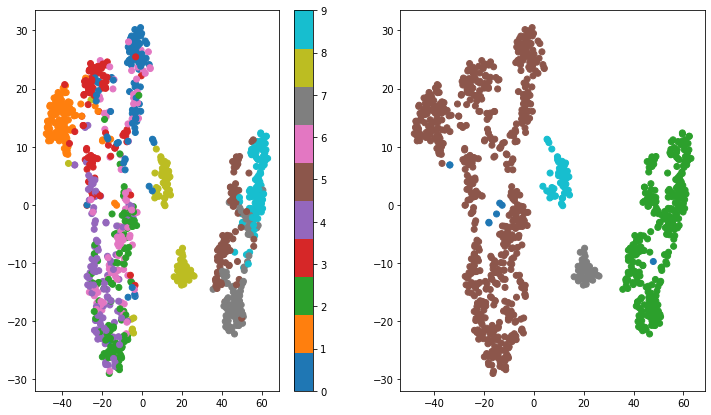}
\caption{\label{FmnistAutoBeforeGradient} Visualisation over true labels (left) and DBSCAN clustering labels (right) in the t-SNE VAE latent space without gradient processing and weight one on reconstruction during training, Fashion-MNIST data.}
\end{figure}

\begin{figure}[!t]
 \centering
\includegraphics[width=250pt, height=200pt]{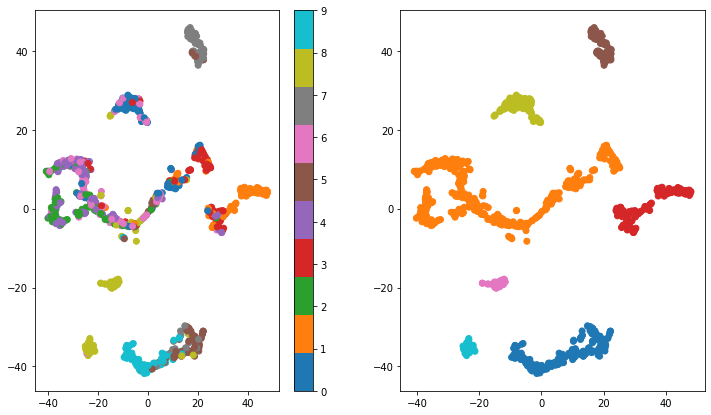}
\caption{\label{FmnistVaeAfterGradient} Visualisation over true labels (left) and DBSCAN clustering labels (right) in the t-SNE VAE latent space after radient processing, Fashion-MNIST data.}
\end{figure}

\begin{figure}[!t]
 \centering
\includegraphics[width=180pt, height=180pt]{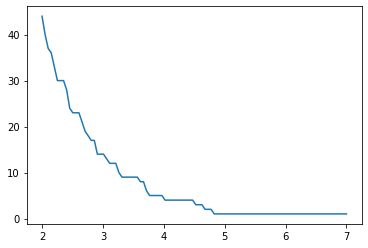}
\caption{\label{FpersistancePre}  Sample persistence graph over t-SNE VAE latent space, without gradient processing, Fashion-MNIST data.}
\end{figure}

\begin{figure}[!t]
 \centering
\includegraphics[width=180pt, height=180pt]{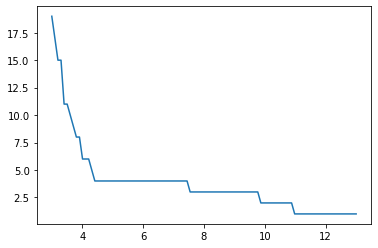}
\caption{\label{FpersistanceAuto}  Sample persistence graph over t-SNE VAE latent space, without gradient processing and weight one on reconstruction during training, Fashion-MNIST data.}
\end{figure}

\begin{figure}[!t]
 \centering
\includegraphics[width=180pt, height=180pt]{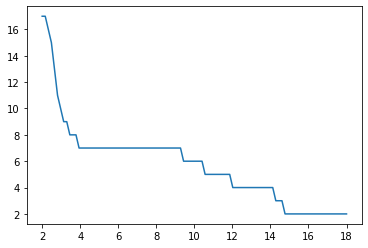}
\caption{\label{FpersistancePost}  Sample persistence graph over t-SNE VAE latent space, with gradient processing, Fashion-MNIST data.}
\end{figure}

\begin{figure}[!t]
 \centering
\includegraphics[width=180pt, height=180pt]{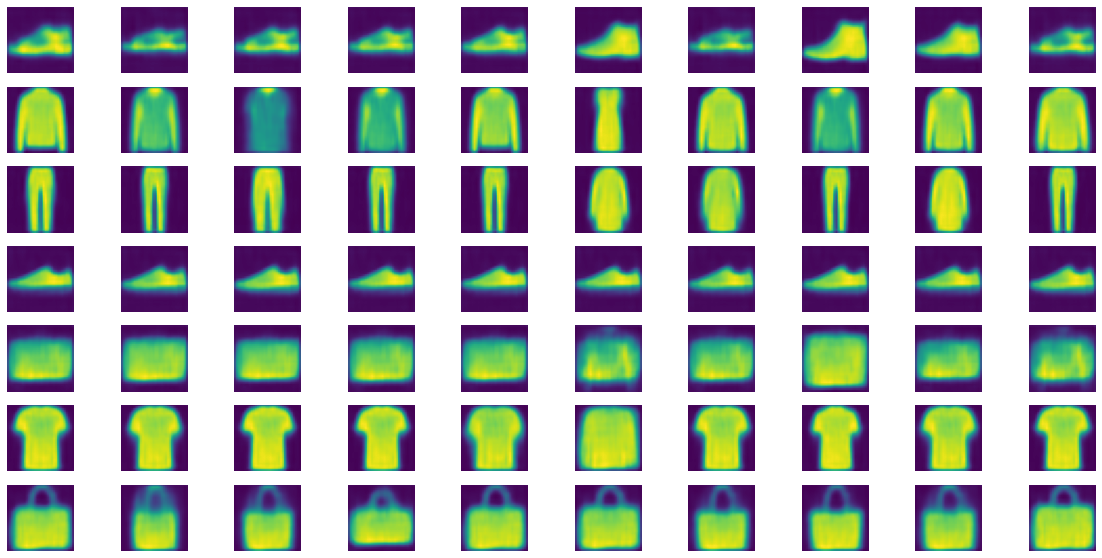}
\caption{\label{ClusterExample}  Samples for each cluster after gradient processing.}
\end{figure}

\begin{figure}[!t]
 \centering
\includegraphics[width=180pt, height=180pt]{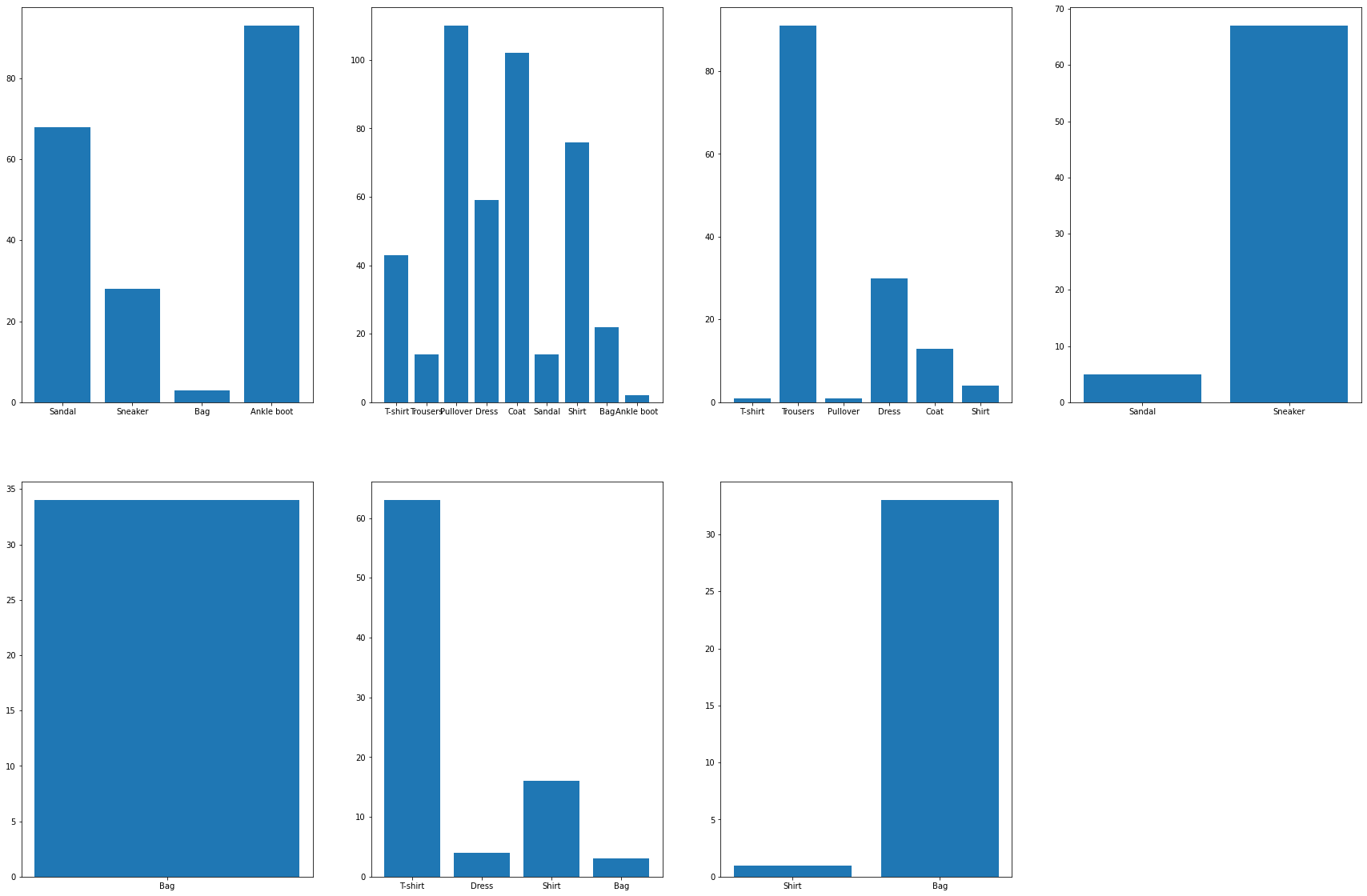}
\caption{\label{ClusterBar}  True label frequency for each cluster after gradient processing.}
\end{figure}

\noindent In this section, the gradient pre-processing of data and cluster counting  algorithm  is  applied  to  the  Fashion-MNIST  data  set\cite{xiao2017}. Fashion-MNIST concist of 28 by 28 grey-scale images of clothes belonging to 10 different classes: T-SHirt, Trouser, Pullover, dress, coat, sandal, shirt sneaker, sneaker, bag and ankle boot. For examples from each class see figure \ref{FmnistExamples}. The setup is nearly identical to the MNIST case in the previous section. A VAE with the same structure as the MNIST case is trained with a latent space dimension of 64. A Autoe.encoder is also trained, can be viewed as a VAE with weight one on the reconstruction during training. In Figure \ref{FmnistVaeBeforeGradient} a t-SNE projection of the latent space for the VAE is shown, both with the true labels and with labels from a DBSCAN clustering with parameter $\epsilon = 2$. The VAE have a hard time distinguishing between the different classes. This is also apparent when looking at the persistence graph corresponding to the VAE, Figure \ref{FpersistancePre}. From the persistence graph no single number of clusters can be detected, again indicating that the VAE does not separate the different classes. Moving on to the Auto-encoder, the t-SNE projection of latent space can be seen in Figure \ref{FmnistAutoBeforeGradient}, both with real labels and with the labels found from a DBSCAN clustering with parameter $\epsilon = 3$. The auto-encoder separates the data better, dividing the data into four distinct clusters. The auto-encoder has grouped all shoes into one group, split Bags into two groups and placed all the other clothes into one big cluster. These four clusters can also clearly be seen in the persistence graph, Figure \ref{FpersistanceAuto}. The persistence graph gives a clear signal of four distinct clusters. Finally the gradient step process is applied to the VAE. A gradient step size of 0.001 is used over $10000$ iterations with a smoothing parameter of $\sigma = 0.0005$. The resulting t-SNE projection of the latent space can be seen in Figure \ref{FmnistVaeAfterGradient}. The data is more clearly separated into distinct clusters. Looking at the persistence graph, Figure \ref{FpersistancePost}, seven number of clusters stands out. To determine hat those seven clusters represent, 10 images from each cluster is shown in Figure \ref{ClusterExample} and the composition of real labels in each cluster is shown in Figure \ref{ClusterBar}. The gradient processing has manege to distinguish two different kind of bags, on with a handle and one without. It has also improved the auto-encoders separation of shoes by distinguishing between sneakers and the rest. Finally for the clothes it has separated t-shirts in its own category, trousers and some dresses in one and placed pullover, coat and shirts in one. These are all semantically meaningful clusters, indicating that the gradient processing step has extracted meaningful information.
\noindent
\subsection{Conclusions}
In this article,  a simple method based on importance sampling for estimating the probability density gradient over data for the VAE framework is presented. It is shown that the local maxima of the  probability density are precisely the points for which reconstruction is perfect on average. We use the gradient to process data, i.e., transport data points by gradient ascent to a local maxima, and demonstrate how to adjust for high number of local maxima by smoothing the objective with a Gaussian kernel. We demonstrate empirically how processing data in this way can improve clustering compared to clustering on unprocessed data. We further present a simple yet effective method for determining a suitable number of clusters for a data set; which is a hyper parameter hard to determine for most advanced clustering methods. The method is based on ideas from persistent homology, and uses the DBSCAN clustering algorithm to efficiently compute the number of connected components of data. It is demonstrated empirically that this cluster counting algorithm performs better on processed data. Interestingly, a baseline autoencoder framework together with t-SNE and GMM (without gradient processing), to our best of knowledge, outperforms all existing clustering methods on MNIST. For future work, we plan to apply this framework on more data sets to investigate its general capabilities to a larger extent.


\begin{thebibliography}{99.}%

\bibitem{Abbasnejad2016}
E.\ Abbasnejad, D.\ Anthony, and A.V.D\ Hengel
\newblock Infinite variational autoencoder for semi-supervised learning.
\newblock \emph{arXiv:1611.07800}, 2016.

\bibitem{Agustsson2019}
E.\ Agustsson, A.\ Sage, R.\ Timofte, L.V\ Gool.
\newblock Optimal transport maps for distribution preserving operations on latent spaces of generative models.
\newblock  \emph{International Conference on Learning Representations}, 2019.

\bibitem{Berthelot2019}
D.\ Berthelot, C.\ Raffel, A.\ Roy and I. Goodfellow.
\newblock Understanding and improving interpolation in autoencoders via adversarial regularizer.
\newblock \emph{ICLR}, 2019.

\bibitem{Ester1996}
M.\ Ester, H.P.\ Kriegel, J.\ Sander and X.\ Xiaowei
\newblock A density-based algorithm for discovering clusters in large spatial databases with noise.
\newblock \emph{Proceedings of the Second International Conference on Knowledge Discovery and Data Mining}, 1996.

\bibitem{Fukunga1975}
K.\ Fukunaga and L.D\ Hostetler
\newblock The estimation of the gradient of a density function, with applications in pattern recognition.
\newblock \emph{IEEE Transactions on Information Theory 21}, 32-40, 1975.

\bibitem{Ghassabeh2015}
Y.A\ Ghassabeh
\newblock A sufficient condition for the convergence of the mean shift algorithm with Gaussian kernel.
\newblock \emph{Journal of Multivariate Analysis 135}, 2015.

\bibitem{Ghassabeh2013}
Y.A\ Ghassabeh
\newblock On the convergence of the mean shift algorithm in the one-dimensional space.
\newblock \emph{Pattern Recognition Letters 34}, 1423–1427, 2013.

\bibitem{Goodfellow2014}
I.\ Goodfellow, J.\ Pouget-Abadie, M.\ Mirza, B.\ Xu, D.\ Warde-Farley, S.\ Ozair, A.\ Courville, and Y.\ Bengio
\newblock Generative Adversarial Networks.
\newblock \emph{Proceedings of the International Conference on Neural Information Processing Systems (NIPS)}, 2672–2680, 2014.

\bibitem{Jiang2017}
Z.\ Jiang, Y.\ Zheng, H.\ Tan, B.\ Tang, and H.\ Zhou.
\newblock Variational deep embedding: An unsupervised and generative approach to
clustering.
\newblock \emph{Proceedings of the Twenty-Sixth International Joint Conference
on Artificial Intelligence, IJCAI}, pages 1965–1972, 2017.

\bibitem{Kingma2013}
D.\ Kingma and M.\ Welling.
\newblock Auto-encoding variational Bayes.
\newblock  \emph{ICLR}, 2014.

\bibitem{Kingma2014B}
D.P\ Kingma, D.J\ Rezende, S.\ Mohamed, and M.\ Welling
\newblock Semi- supervised learning with deep generative models.
\newblock \emph{NIPS}, 2014.

\bibitem{Kulczycki2010}
P.\ Kulczycki and M.\ Charytanowicz
\newblock A complete gradient clustering algorithm formed with kernel
estimators.
\newblock \emph{International Journal of Applied Mathematics and Computer Science 20}, 123-134, 2010.

\bibitem{LeCun1998}
Y.\ LeCun, L.\ Bottou, Y.\ Bengio, and P. Haffner.
\newblock Gradient-based learning applied  to  document recognition.
\newblock \emph{Proceedings of the IEEE}, 86(11):2278-2324, 1998.

\bibitem{Lewis2004}
D.\ Lewis, Y.\ Yang, T.G\ Rose, and F.\ Li.
\newblock Rcv1: A new benchmark collection for text categorization research.
\newblock \emph{Journal of machine learning research}, 2004.

\bibitem{Maaloe2016}
L.\ Maaløe, C.K\ Sønderby, S.K\ Sønderby, and O.\ Winther
\newblock Auxiliary deep generative models.
\newblock \emph{ICML}, 2016.

\bibitem{Maaten2008}
L.J.P\ van der Maaten and G.E\ Hinton.
\newblock Visualizing Data Using t-SNE.
\newblock \emph{Journal of Machine Learning Research}, 2008.

\bibitem{Makhzani2016}
A.\ Makhzani, J.\ Shlens, N.\ Jaitly, I.\ Goodfellow and B.\ Frey
\newblock Adversarial autoencoders.
\newblock \emph{NIPS}, 2016.

\bibitem{Makhzani2017}
A.\ Makhzani and B.J\ Frey
\newblock Pixelgan autoencoders.
\newblock \emph{Advances in Neural Information Processing Systems 30: Annual Conference on Neural Information Processing Systems}, 1972–1982, 2017.

\bibitem{Nalisnick2016}
E.\ Nalisnick and P.\ Smyth
\newblock Stick-breaking variational autoencoders.
\newblock \emph{arXiv:1605.06197}, 2016.

\bibitem{CarlAdamHenrikCode}
C.\ Ringqvist and A.\ Lindhe
\newblock 1aa4sgNhbDmNfGDK8wHhPcSs2eXo1Hrv4
\newblock \emph{Codebase for paper hosted at https://colab.research.google.com/drive/}, 2021.

\bibitem{Salimans2016}
T.\ Salimans, I.\ Goodfellow, W.\ Zaremba, V.\ Cheung, A.\ Radford, and X.\ Chen
\newblock Improved techniques for training gans.
\newblock \emph{NIPS}, 2016.

\bibitem{Unlu2019}
R.\ \"Unl\"u and P.\ Xanthopoulos  \newblock Estimating the number of clusters in a dataset via consensus clustering. 
\newblock \emph{Expert Systems with Applications}, 125(1), 33-39,  2019.

\bibitem{xiao2017}
H.\ Xiao, K.\ Rasul, R.\ Vollgraf
\newblock Fashion-MNIST: a Novel Image Dataset for Benchmarking Machine Learning Algorithms
\newblock \emph{arXiv} cs.LG/1708.07747 

\bibitem{Xie2016}
J.\ Xie, R.\ Girshick, and A.\ Farhadi.
\newblock Unsupervised deep embedding for clustering analysis.
\newblock \emph{ICML}, 2016.

\bibitem{Zhao2019}
W.\ Zhao, S.\ Wang, Z.\ Xie, J.\ Shi and C.\ Xu
\newblock GAN-EM: GAN based EM Learning Framework.
\newblock \emph{International Joint Conference on Artificial Intelligence (IJCAI)}, 2019.





\end{thebibliography}
\end{document}